\documentclass[conference]{IEEEtran}

\usepackage{graphicx}
\usepackage{amsmath}
\usepackage{cleveref}
\usepackage{algorithmic}
\usepackage[ruled,vlined]{algorithm2e}
\usepackage{xcolor}
\usepackage{nicefrac}
\usepackage{caption}
\usepackage{subcaption}
\usepackage{booktabs}
\usepackage{cancel}
\usepackage{cite}
\usepackage{multirow}
\usepackage{siunitx}
\usepackage{ulem}

\usepackage{enumitem}

\newlist{inlinelist}{enumerate*}{1}
\setlist*[inlinelist,1]{%
  label=\arabic*),
}

\newlist{inlineitemlist}{enumerate*}{1}
\setlist*[inlineitemlist,1]{%
  label=\null,
}

\begin{document}

\title{Efficient automated U-Net based tree crown delineation using UAV multi-spectral imagery on embedded devices}

 \author{%
     \IEEEauthorblockN{Kostas~Blekos\IEEEauthorrefmark{1}, Stavros~Nousias\IEEEauthorrefmark{1}, Aris~S~Lalos\IEEEauthorrefmark{1}}
     \IEEEauthorblockA{
         \IEEEauthorrefmark{1} Industrial Systems Institute, ATHENA Research Center, Patras, Greece
    }
     \{blekos,nousias,lalos\}@isi.gr
 \vspace{-1.5em}
 }
\maketitle

\begin{abstract}
Delineation approaches provide significant benefits to various domains, including agriculture, environmental and natural disasters monitoring. 
Most of the work in the literature utilize traditional segmentation methods that require a large amount of computational and storage resources. 
Deep learning has transformed computer vision and dramatically improved machine translation, 
though it requires massive dataset for training and significant resources for inference. 
More importantly, energy-efficient embedded vision hardware delivering real-time and robust performance is crucial in the aforementioned application.  
In this work, we propose a U-Net based tree delineation method, which is effectively trained using multi-spectral imagery but can then delineate single-spectrum images. 
The deep architecture that also performs localization, i.e., a class label corresponds to each pixel, has been successfully used to allow training with \color{black}
a small set of segmented images.
The ground truth data were generated using traditional image denoising and segmentation approaches. To be able to execute the proposed DNN efficiently in embedded platforms designed for deep learning approaches, we employ traditional model compression and acceleration methods. Extensive evaluation studies using data collected from UAVs equipped with multi-spectral cameras demonstrate the effectiveness of the proposed methods in terms of delineation accuracy and execution efficiency.
\color{black}
\end{abstract}

\begin{IEEEkeywords}
CNN, Accelerated CNN, Deep learning, U-net, Image segmentation
\end{IEEEkeywords}

\IEEEpeerreviewmaketitle

\section{Introduction}
Remote sensing is an important tool in automated and precision agriculture, forestry inspection and management. In the agricultural sector, remote sensing facilitates soil management, decease and weed detection\cite{thorp2004review}, pest management, evaluation of vegetation health and vigor, among other needs. Tree crown delineation isolates regions of interest for extracting vegetation indices locally, providing higher resolution ranges that facilitate the accurate detection of potential diseases linked to water stress.
Multiple studies address the problem using high-resolution RGB images \cite{gougeon1995crown} or multi-spectral intensity imaging data featuring characteristic bands of the light spectrum~\cite{brandtberg1998automated,yang2014multi,li2019real}.
Normalized difference vegetation index (NDVI) employs red and near-infrared to characterize vegetation \cite{weier2000measuring}.
Light detection and ranging (LIDAR) data \cite{zhao2007hierarchical,weinstein2019individual} are also employed. All the remote sensing data originate from unmanned aerial vehicles \cite{huang2018individual,santos2019assessment}, drones or satellites \cite{gomes2018individual}. 
Several algorithms are present in the literature addressing the problem of tree detection and delineation for feature extraction and segmentation. 
Current tree segmentation approaches are primarily based on user-defined algorithms that describe the appearance of trees in a hierarchical sequence of rules.  
These approaches may be broadly categorized~\cite{ke2008comparison,ke2011review} as either local maxima-minima \cite{li2019real}, contour detection \cite{ke2010active,lin2011multi}, region growing \cite{zhen2015agent,dalponte2019individual}, template matching, \cite{dai2018new} valley following \cite{gougeon1995crown}, edge detection approaches \cite{brandtberg1998automated} and watershed routines \cite{zhao2007hierarchical,jing2012individual,yang2014multi,duncanson2014efficient}. 
A recent comparative study \cite{aubry2019comparative} suggests that there is no definitive approach given the variety of forest formations and species, concluding that crown segmentation in a multi-layered closed canopy is significantly improved using 3D segmentation from LIDAR data than relying on the surface RGB images. Further improvements are expected when combining them.

Deep learning and learning-based approaches \cite{li2019real,weinstein2019individual,csillik2018identification} also exhibit commendable results. 
A recent semi-supervised approach \cite{weinstein2019individual}, employing a convolutional neural network (CNN),  combines LIDAR and RGB data, yielding similar outcomes with classical unsupervised algorithms. CNNs were also used with multi-spectral imaging data \cite{csillik2018identification,santos2019assessment}. In \cite{csillik2018identification}, a deep network was employed to differentiate trees, bare soil and weeds. Li et al. \cite{li2017deep} developed a CNN framework to detect oil palm trees. Even though they provide accurate results, they need a large amount of training data. Training should be performed in a sliding window setup to predict the class of each pixel. The process receives as input a local neighbourhood. As a result, the training set is much larger than the number of images. The process is slow since it has to be executed for each patch separately. 
A trade-off comes at play concerning the patch size. Larger patches tend to require more max-pooling layers that reduce the accuracy, while small patches miss the correlation between features and context information. 
To this end, variants of a more elegant scheme, the fully convolutional network (FCN), need to be employed. 
We utilize the U-Net, a modified version of the FCNs where high-resolution features from the contracting path are combined with the expanding path. 
A typical convolutional neural network is followed by a series of convolutional layers where the pooling operators are replaced by upsampling operators increasing the resolution. Feature maps from the downsampling path are concatenated with the upsampled output allowing for the network to localize. As a result, a symmetric U-shaped architecture is formed. The U-Net allows for training with relatively small dataset while generating more precise segmentations as the authors in \cite{ronneberger2015u} highlight. 
Motivated by the aforementioned open issues and challenges, we propose a U-Net based tree delineation method, using multi-spectral imagery. 
We generate a small groundtruth dataset using traditional image preprocessing and segmentation approaches, effectively training the deep network. 
Several acceleration approaches were employed, facilitating the deployment on Edge-TPU devices mounted on UAV and drone chassis.
More specifically, the contributions of the proposed approach can be summarized in the following points: \begin{inlinelist}
    \item We generate groundtruth segmentation data using as input multi-spectral imagery ranging in four light spectrum bands
    \item We denoise, align and process the multi-spectral images 
    to generate groundtruth data.
    \item \color{black} We accurately perform tree crown detection even when we use the data from a single band, using a U-Net based architecture trained with the generated segmentation masks.
    \color{black}
    \item We present qualitative, quantitative and performance evaluation of the presented approaches.
    \item We implemented the presented approaches to be executed on a Google Coral Edge.
\end{inlinelist}
The rest of this paper is organized as follows: Section~\ref{sec:method} analyses the methodology, section~\ref{sec:results} presents the results of our approach, while conclusions are drawn in Section~\ref{sec:conclusion}.

\begin{figure}
    \centering
\begin{subfigure}{0.49\linewidth}
  \includegraphics[width=\linewidth]{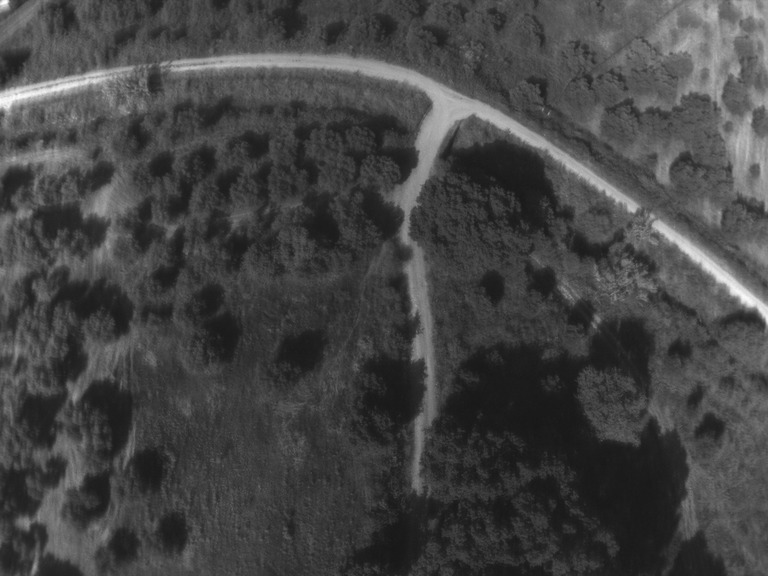}
  \caption{}
  \label{fig:spectrum_GRE}
\end{subfigure}
\begin{subfigure}{0.49\linewidth}
  \includegraphics[width=\linewidth]{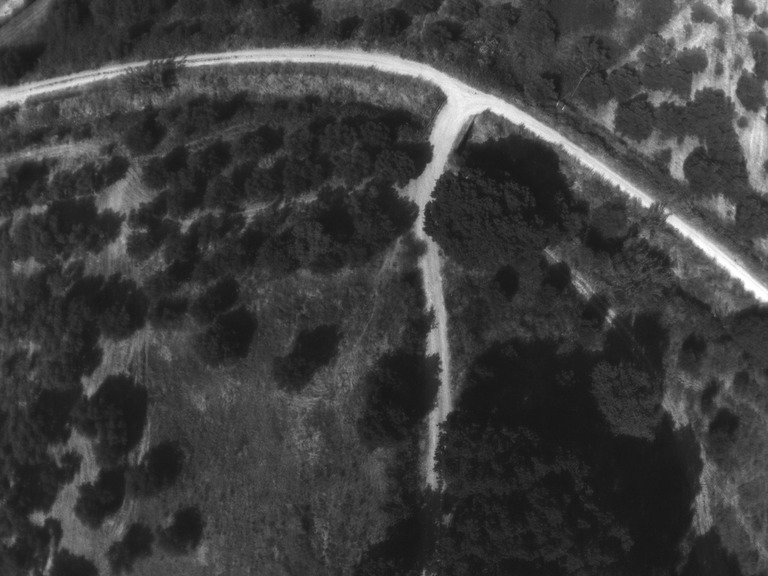}
  \caption{}
  \label{fig:spectrum_RED}
\end{subfigure}
\begin{subfigure}{0.49\linewidth}
  \includegraphics[width=\linewidth]{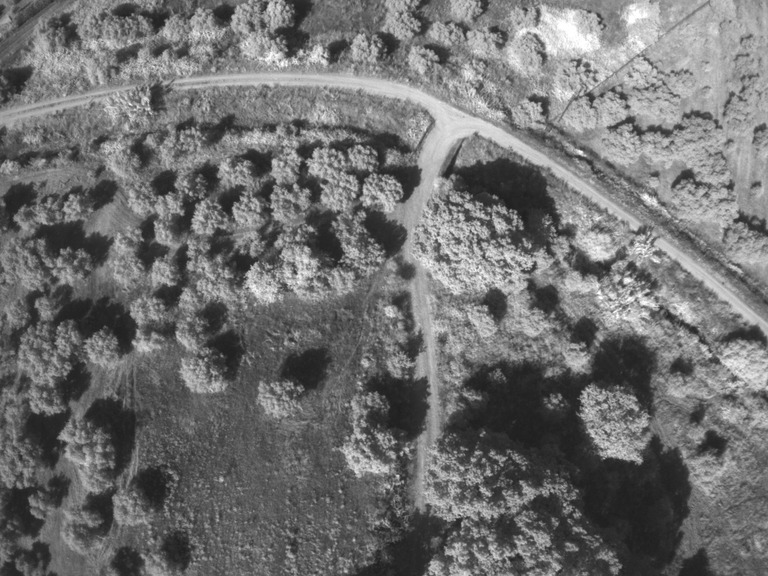}
  \caption{}
  \label{fig:spectrum_REG}
\end{subfigure}
\begin{subfigure}{0.49\linewidth}
  \includegraphics[width=\linewidth]{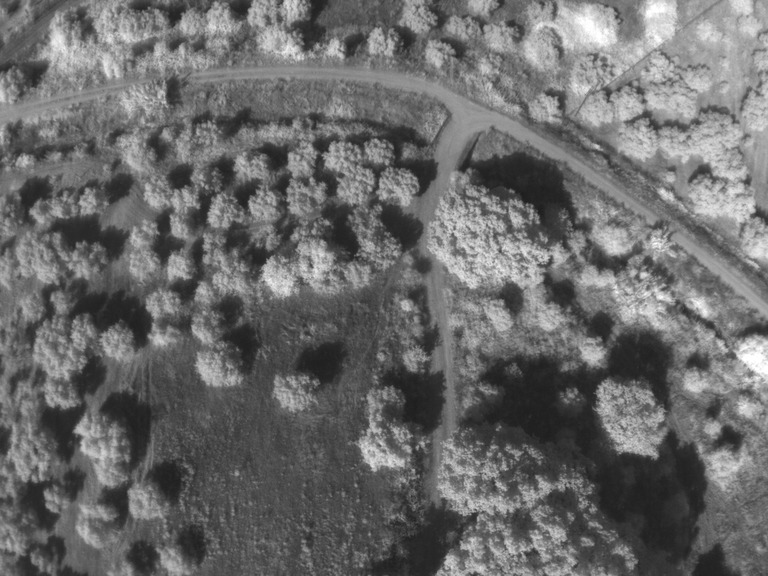}
  \caption{}
  \label{fig:spectrum_NIR}
\end{subfigure}
    \caption{ \small
    Sample of raw input images. One image per spectral band, taken at the same time using a multispectral camera. (a) GRE - Green 550nm (b) RED - Red 660nm (c) REG - Red Edge 735nm (d) NIR - Near Infrared 790nm}
    \label{fig:bands}
    \vspace{-1em}
\end{figure}


\section{Deep architectures for tree crown delineation}\label{sec:method}

\begin{figure}[t]
\includegraphics[width=\linewidth]{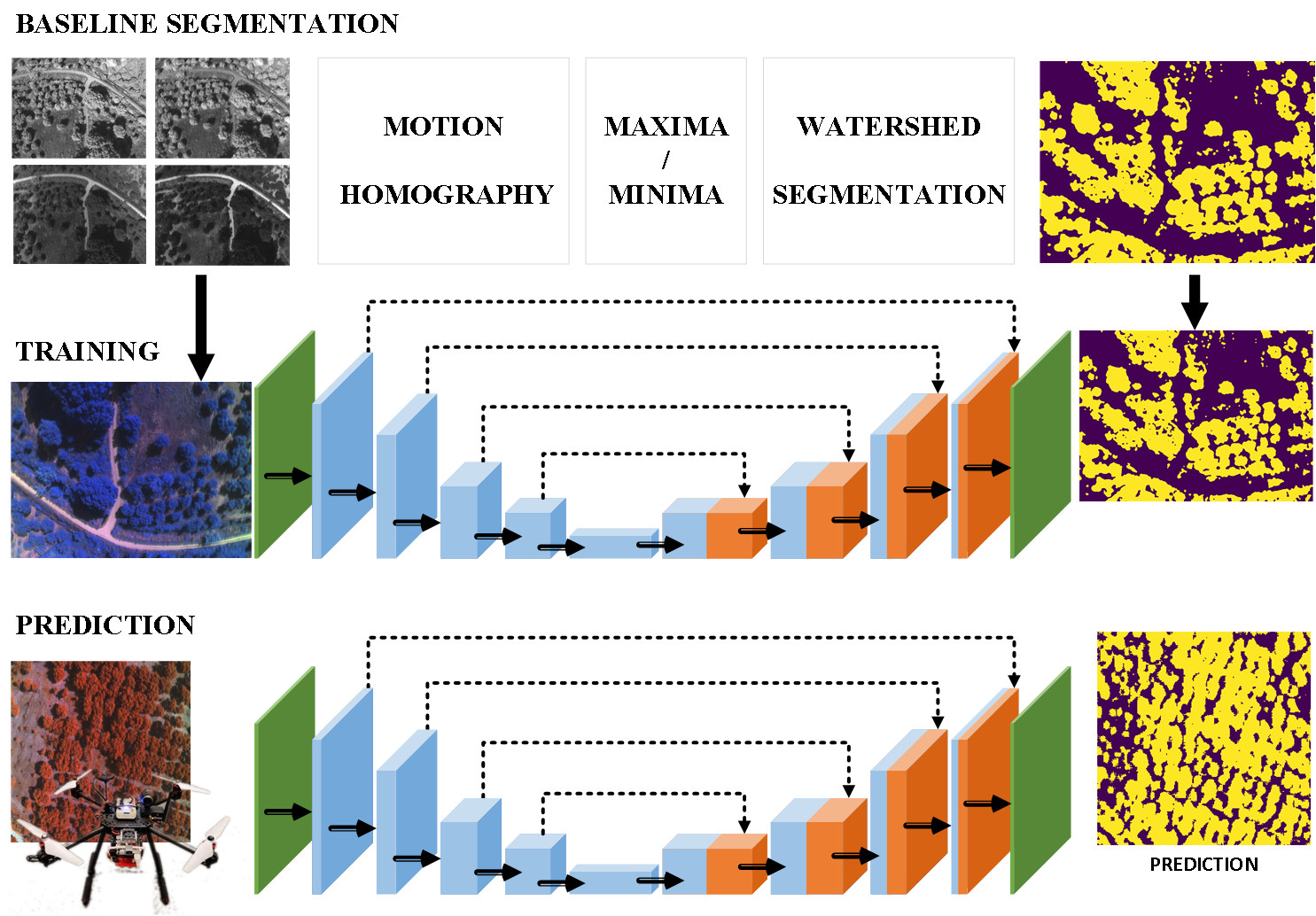}
\caption{\small Processing pipeline}
\label{fig:pipeline}
\vspace{-1em}
\end{figure}

This section analyzes the processing pipeline of our approach. 
Subsection \ref{subsection:classical} describes the extraction of tree segmentation masks. 
The generated masks are employed to train a U-Net that is subsequently deployed on Google Coral Edge devices mount on UAVs and drone frames. An overview of the architecture and training process of the proposed deep network is depicted in Figure \ref{fig:pipeline}.
\begin{figure}
    \centering
\begin{subfigure}{\linewidth}
\centering
  \includegraphics[width=\linewidth]{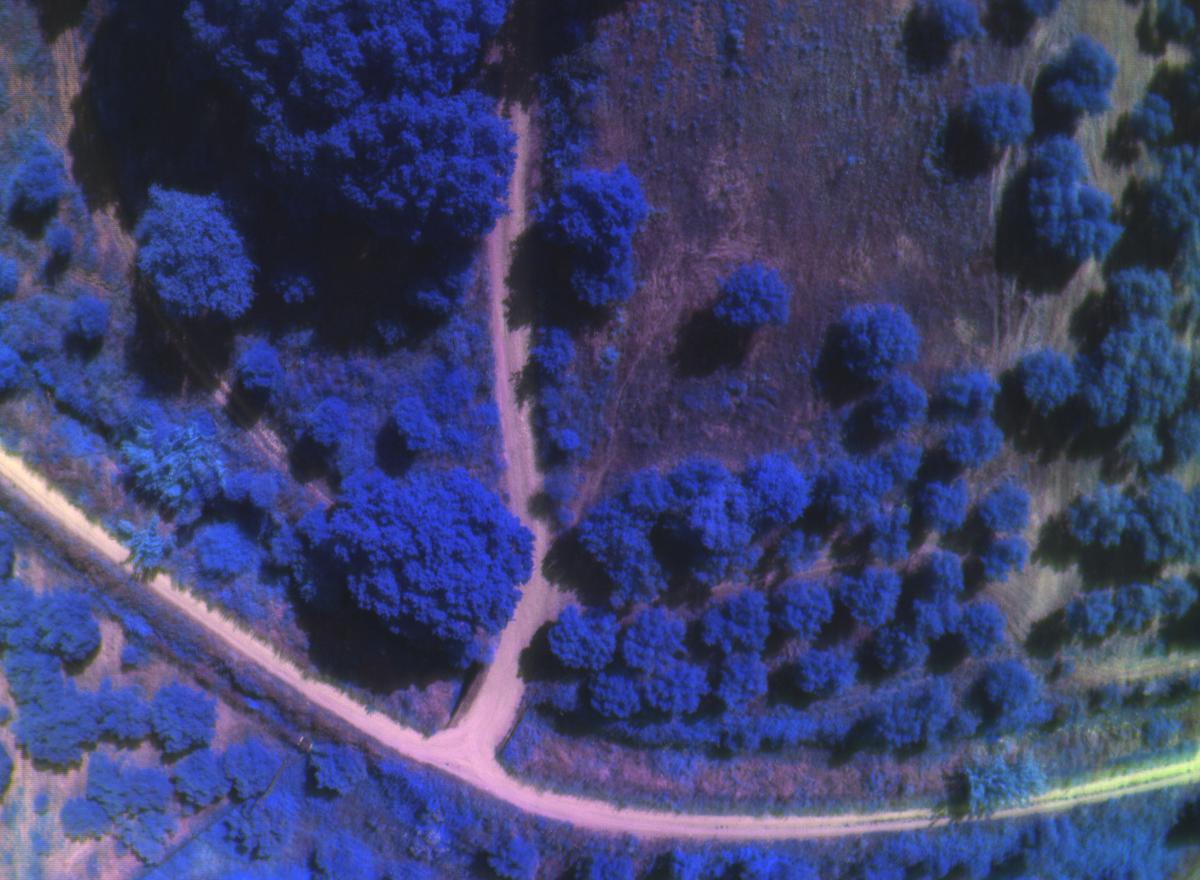}
  \caption{}
  \vspace{0.5em}
  \label{fig:segmentation_1}
\end{subfigure}
\begin{subfigure}{0.49\linewidth}
  \includegraphics[width=\linewidth]{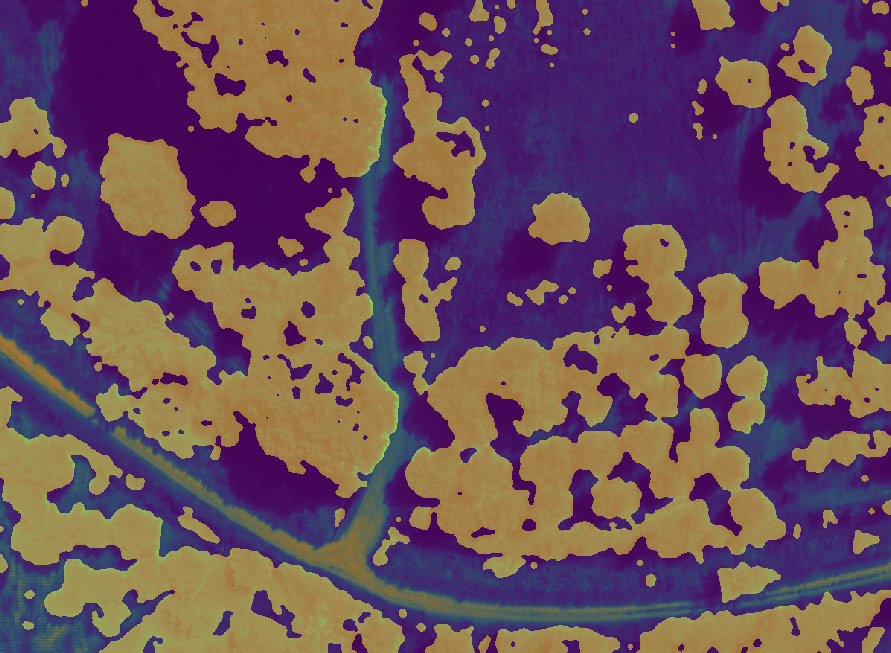}
  \caption{}
  \label{fig:segmentation_2}
\end{subfigure}
\begin{subfigure}{0.49\linewidth}
  \includegraphics[width=\linewidth]{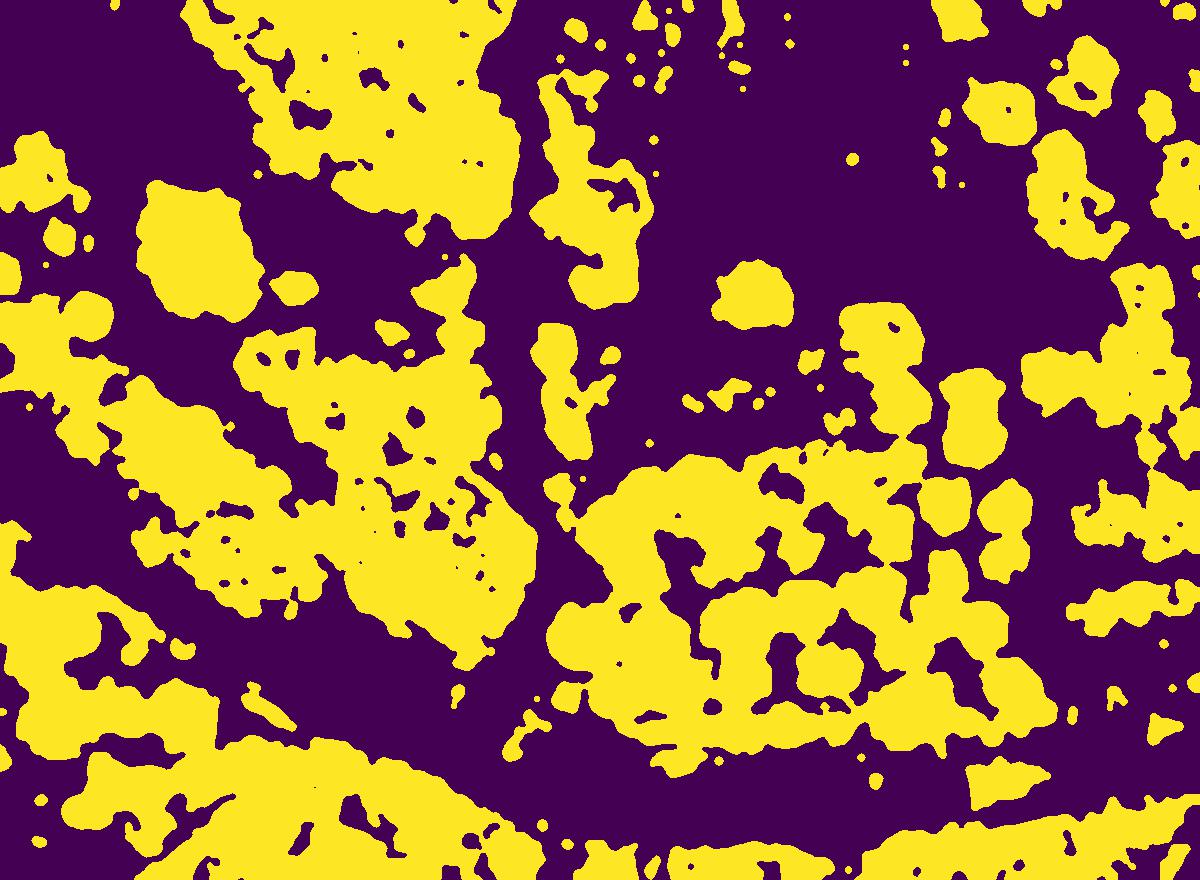}
  \caption{}
  \label{fig:segmentation_3}
\end{subfigure}
    \caption{ \small Baseline segmentation for an input image with road. Composite pseudo-RGB input (a), overlay of input and ground truth (b), automatically generated ground truth labels (c). }
    \label{fig:groundtruth_overlay_1}
    \vspace{-1em}
\end{figure}


\subsection{Tree characterization through analysis of multispectral imagery}\label{subsection:classical}

\paragraph*{Band alignment}
As input for the primary step of our approach, multi-spectral intensity images are used.
The green channel corresponds to 550nm, referred to as GRE, 
the red channel to 660nm referred to as RED, 
the red edge channel to 735nm referred to as REG.
Finally, the near-infrared channel corresponds to 790nm, referred to as NIR. 
Initially, the four channels are unregistered (see figure~\ref{fig:bands}) so
we employ simple motion homography \cite{Hartley2004} to align the images.

Usually, image registration is not straight forward and requires many resources,
especially when, as in our case,
there are significant differences
between the input images.
To efficiently and accurately align them, 
we process them successively in pairs, based on the similarity
of the spectral response of each band: (RED to GRE), (GRE to REG) and finally (REG to NIR). 
Combining successive pairing with motion homography yields satisfactory results.

\paragraph*{Composite image creation}
After alignment, the unmatched regions are cropped
and a 4-dimensional composite image is created.
Another composite pseudo-RGB image is formulated,
to emulate RGB image input.
This pseudo-RGB images 
uses the GRE, RED and REG bands, 
and is used as training input to the neural network.

\paragraph*{Distance calculation}
After the above preprocessing steps,
we proceed to generate the ground truth labels.
The first step is to calculate the distance of each pixel
of the composite image from the reference point.
The reference point (RP) is the point in the multi-spectral intensity space (GRE, RED, REG, NIR)
that shows the maximum correlation with the spectral response of the trees to be delineated.
This point can be extracted from spectral data available from the literature,
or experimentally estimated from the input data.
In this work, we used the RED-normalized RP (1.29, 1.00, 3.13, 2.76).

\paragraph*{Local minimum filtering}
Subsequently, a minimum filter is applied to mark regions that contain a tree
with high probability, as regions that \emph{contain} a point very close to the RP
have a very high probability of belonging to a tree. 
Roughly, the probability that the center pixel of a region $A$ of area $S_A$ belongs to a tree is given by
\begin{equation}
    P(A) \propto \frac 1{S_A}\frac 1{1+\min_{p_i\in A}\left\lVert p_i,p_{\mathrm{RP}}\right\rVert}
\end{equation}
so a minimum filter is a good choice for marking tree regions.

\paragraph*{Denoising}
After applying the local minimum filter, a median filter, of disc-shaped kernel size $k$, is applied to remove unneeded details and smooth out the outcome. 
The kernel size is correlated to the expected or the desirable minimum radius of tree features.
After experimentation with kernel sizes,
a kernel of $k=5$ pixels, corresponding to a real length of about $50$cm, was found to be adequate.

\paragraph*{Thresholding}
Markers are identified at locations that the probability is lower than an experimentally defined threshold $\theta_m$.
The best results were found for $\theta_m = 0.15$, but results around that number showed little variation,
as most pixels are clearly in one class or the other; the in-between values are outliers and represent
a small percentage of the image, the most important of which are around the edge of the tree crown
or in shaded areas.

\paragraph*{Watershed segmentation}
Finally, a watershed algorithm \cite{grau2004improved} is employed for the final segmentation outcome,
to highlight the tree crown.

The outcome of this procedure is presented in Figure \ref{fig:groundtruth_overlay_1}.
Algorithm \ref{alg:baselinesegmentation} highlights the processing steps.

\begin{algorithm}
\SetAlgoLined
\KwResult{Per pixel classification: 0 - not a tree, 1 - tree }
 Band alignment\;
 Composite image creation\;
 Distance calculation\;
 Local minimum filtering\;
 Denoising\;
 Threshold classification\;
 Watershed segmentation\;
 \caption{Generation of ground truth labeled images.}
 \label{alg:baselinesegmentation}
\end{algorithm}

\begin{figure}
    \centering
    \includegraphics[width=\linewidth]{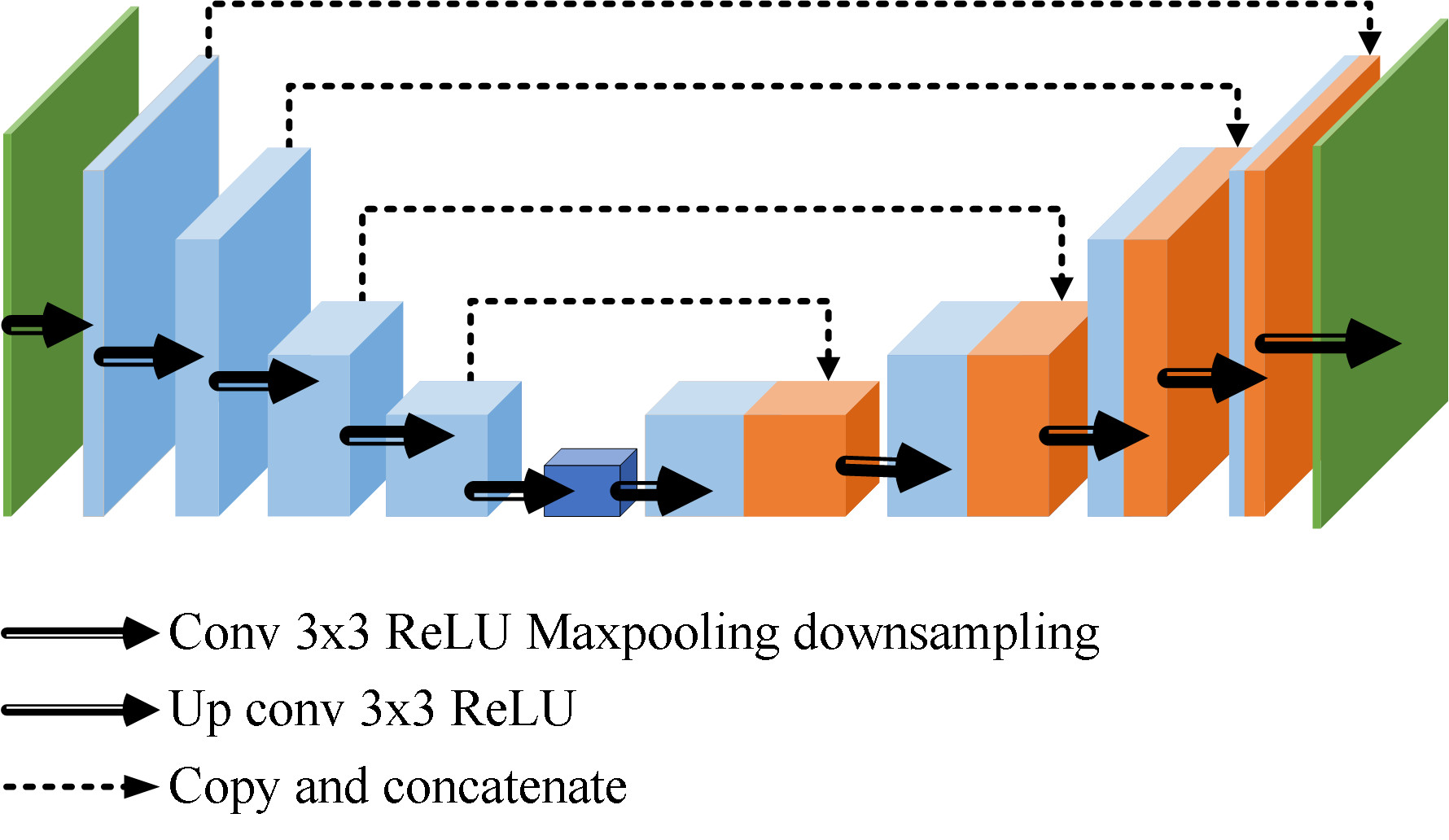}
\caption{\small U-Net architecture}
\label{unet_architecture}    
\vspace{-1em}
\end{figure}

\subsection{U-Net architecture, training and inference}
\label{subsection:unet}

Deep architectures are able to learn complex non-linear relationships found in the data and, in our case, to translate the input image to a segmentation map. 
The U-Net architecture \cite{ronneberger2015u} builds upon the concept of fully connected convolutional neural networks and has been successfully employed to segment biomedical images. 
The motivation behind fully connected CNNs is to accelerate the segmentation process so as to be executed on edge devices. 
In the U-Net architecture (Figure \ref{unet_architecture})
the input data are fed to successive down-scaling layers (left branch) to reduce the spatial resolution of the feature maps and then to corresponding up-scaling layers (right branch), increasing the spatial resolution of the feature maps.
Each layer consists of two 3x3 unpadded convolutions succeeded by a rectified linear unit (ReLU) activation function, a 2x2 max pooling with stride 2. After each down-sampling, the number of feature channels equals twice the number of channels of the previous layer.
In the up-scaling part, every layer performs of a 2x2 convolution. The number of feature channels equals half the number of channels of the previous layer. The result is concatenated with the corresponding part of the down-scaling path.  Finally, two 3x3 convolutions are performed, succeeded by a ReLU. The cropping is necessary due to the loss of border pixels in every convolution. At the final layer, a 1x1 convolution is used to map each 64- component feature vector to the desired number of classes. At the final layer, a 1x1 convolution is used to map each feature vector to the desired number of classes. In total the network has 23 convolutional layers. The outcomes of the primary baseline segmentation step are used as input for the training of the U-Net architecture.
Each training example consists of the composite pseudo-RGB image and the corresponding segmentation map. A composite image is used as input for the deep architecture, referred to as multi-spectral setup for the rest of the paper. A second setup involves only the GRE channel to be used as input data. The latter will be referred to as the one-band setup for the rest of the paper. In the following sections, we present a qualitative, a quantitative and performance evaluation for both strategies in terms of execution times. 

\subsection{Squeezing the proposed U-Net architecture}
Post-training quantization is an important step to reduce requirements in CPU, processing power, and model size with little degradation in model accuracy. Quantization is performed on an already-trained float TensorFlow model and applied during TensorFlow Lite conversion facilitating the execution of the trained U-Net on the Edge-TPU. The model parameters are quantized to 8bit integers. 
The motivation is that such simplifications facilitate the real-time condition with acceptable losses. 
We create the Quantized FlatBuffer format
using a TFLite Converter object from the saved Keras model and set the inference output type to be an unsigned 8-bit integer. The converter uses images from the dataset to calibrate the model on inputs. Finally, the saved model is compiled as an Edge-TPU executable. 




\section{Experimental evaluation}\label{sec:results}

\subsection{Dataset acquisition and training}
This subsection describes the dataset collection and the training process. 
We present two different setups, a multi-spectral setup and a one-band setup, and we investigate the effect of dataset size on the reported accuracy, 
to demonstrate that the U-Net exhibits high accuracy even with small datasets.

\subsubsection{Dataset} We created a dataset from multi-spectral images collected with Pix4d Parrot Sequoia cameras attached on a C0 class drone. 
The drone collected 400 images on 4 flight hours over olive groves located in western Greece. 
An example of the collected images is presented in Figure \ref{fig:bands}. 

\subsubsection{Multi-spectral setup} 
For the training of the U-Net pairs of multi-band images and ground truth segmentation maps are utilized. 
The multi-band images are generated by the process presented in subsection \ref{subsection:classical}. 
In total, 85 random pairs were used, randomly split in training and test sets at a ratio 80\%-20\%. 
Sparse categorical cross-entropy was used as a cost function with Adam optimizer with a learning rate of $10^{-3}$. 
Training took place for 70 epochs in an in an Intel(R) Core(TM) i7-7600U CPU @ 2.80GHz with 16GB RAM utilizing 4 cores.

\subsubsection{One-band setup}
A second training setup takes place with pairs of one-band images and ground truth segmentation maps. The motivation behind utilizing the one-band setup is to lower the preprocessing execution time and facilitate the execution in edge devices and allow for real-time inference. 

\subsubsection{Size of training dataset and number of epochs}
\begin{figure}
    \centering
    \begin{subfigure}{0.48\linewidth}
      \includegraphics[width=\linewidth]{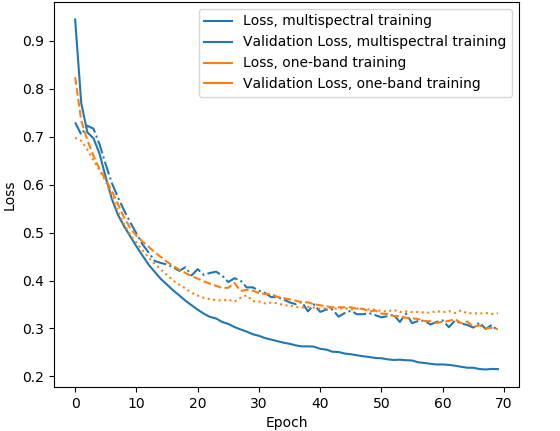}
      \caption{}
    \end{subfigure}
    \begin{subfigure}{0.48\linewidth}
      \includegraphics[width=\linewidth]{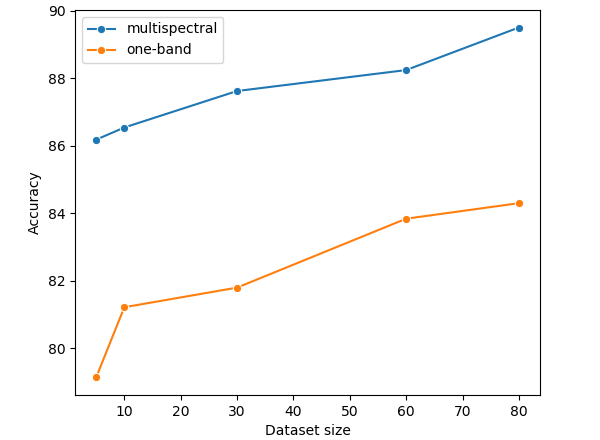}
      \caption{}
    \end{subfigure}
    \caption{\small (a) Training and validation loss for the two training setups: multispectral and one-band. (b) Accuracy as tested on the extended (100 images) dataset vs training data set size (average over 5 training sessions for each set).}
    \label{fig:loss}
    \vspace{-1em}
\end{figure}

We investigated several training setups evaluating the test accuracy with respect to a different number of epochs and sizes for the training dataset. 
Figure \ref{fig:loss}-a presents the training loss and test accuracy for different dataset size. 
Figure \ref{fig:loss}-b demonstrates that U-Net allows for even small datasets to yield relatively high accuracy. 

\subsubsection{Acceleration approaches}
Several adaptations were taken into account for deployment to Edge-TPU. 
The resulting model was converted using TensorFlow Lite to an optimized FlatBuffer format, without quantization, and a Quantized FlatBuffer format. 
The Quantized model was then compiled for use on Edge-TPU devices using the \texttt{edgetpu\_compiler}.
The resulting FlatBuffer model was nearly 0.25 times the size of the base model,
while the Quantized model was 0.1 times the size of the original model as presented in Table \ref{table:results}. 
For the evaluation, two setups were deployed: i) An Intel(R) Core(TM) i7-7600U CPU @ 2.80GHz with 16GB RAM 
and ii) A Google  Coral  DevBoard Edge-TPU device with quad  Cortex-A53, Cortex-M4F  CPU,  Integrated GC7000 Lite Graphics, Google Edge TPU coprocessor as an ML accelerator, 1 GB LPDDR4 main memory.


\begin{table*}
\caption{Comparison of trained models in terms of execution times (inference) and accuracy}

\resizebox{\textwidth}{!}{  

\begin{tabular}{|l|c|c|c|c|}
\hline
\multirow{3}{*}{\textbf{Model}}                  
& \multicolumn{2}{c|}{\textbf{Inference times measused in ms}}                                         & \multicolumn{2}{c|}{\multirow{2}{*}{\textbf{Accuracy (\%)}}}                                           \\ \cline{2-3}
                                        & \textbf{Intel(R) Core(TM)}        & \textbf{Google Coral} & \multicolumn{2}{c|}{}                         \\ \cline{2-5} 
                                        & \textbf{i7-7600U CPU  @  2.80GHz} & \textbf{Edge TPU}     & \textbf{Multispectral input} & \textbf{GRE input band only} \\ \hline
\textbf{Base model}     & 96                                                 &                                        & 89                                            & 84                                            \\ \hline
\textbf{Quantized-CPU} & 5600                                               & 740                                    & \multirow{2}{*}{88}                           & \multirow{2}{*}{83}                           \\ \cline{1-3}
\textbf{Quantized-TPU} &                                                    & 28                                     &                                               &                                               \\ \hline
\end{tabular}

}
\label{table:results}
\vspace{-1em}
\end{table*}


\subsection{Results}

\begin{table}
    \caption{Total execution time}
     \centering
     \begin{tabular}{|l||l|c|c|}
\hline
                                    & \textbf{Operation}                         & \multicolumn{2}{c|}{\textbf{Time (ms)}}     \\ 
                                    \hline \hline
\multirow{3}{*}{\textbf{Training}}  & \multirow{2}{*}{Preprocessing (per image)} & \multicolumn{2}{c|}{\multirow{2}{*}{5450}}  \\
                                    &                                            & \multicolumn{2}{c|}{}                       \\ \cline{2-4} 
                                    & Training (per epoch)                       & \multicolumn{2}{c|}{12200}                  \\ \hline \hline
\multirow{4}{*}{\textbf{Inference}} &                                            & \textbf{One band} & \textbf{Multi spectral} \\ \cline{2-4} 
                                    & Preprocessing (per image)                  & 110               & 3920                    \\ \cline{2-4} 
                                    & I/O operations                             & \multicolumn{2}{c|}{20}                     \\ \cline{2-4} 
                                    & Inference                                  & \multicolumn{2}{c|}{Table I}                \\ \hline
\end{tabular}
 
     \label{tab:execution_time}
     \vspace{-1em}
 \end{table}


\begin{figure}
    \centering
    \begin{subfigure}{0.48\linewidth}
      \includegraphics[width=\linewidth]{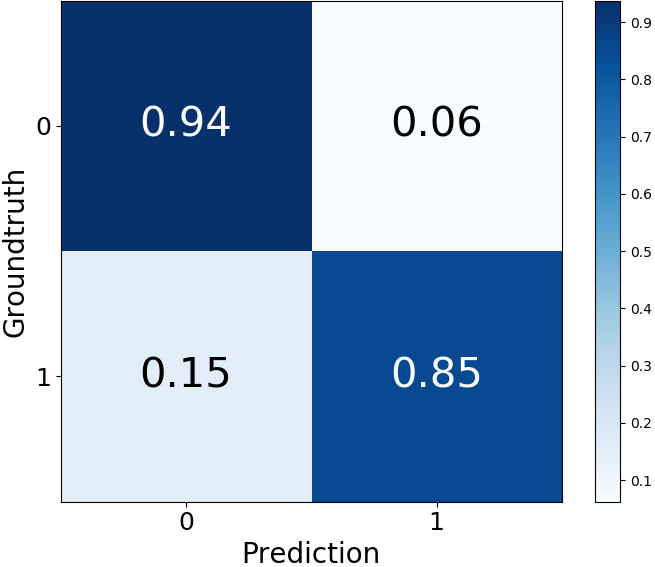}
      \caption{}
    \end{subfigure}
    \begin{subfigure}{0.48\linewidth}
      \includegraphics[width=\linewidth]{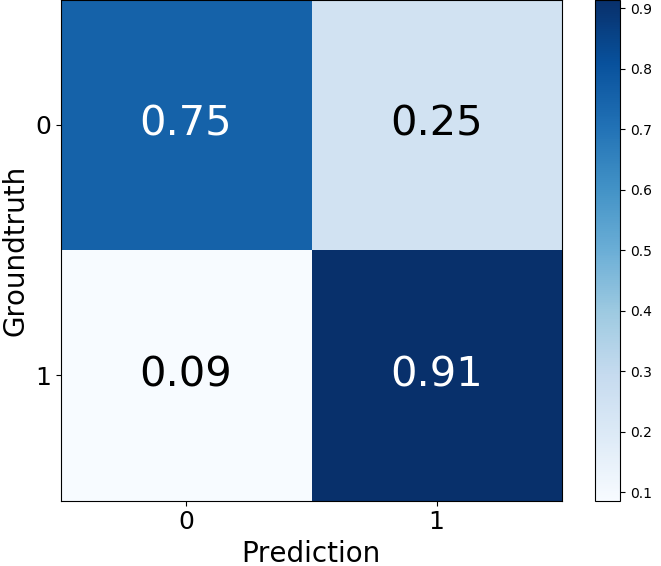}
      \caption{}
    \end{subfigure}

\caption{\small (a) Confusion matrix for multi-band, reported accuracy 89\%. (b) Confusion matrix for the one-band set, reported accuracy 84\%.}
\label{fig:confusion_multi}
\vspace{-1em}
\end{figure}



\begin{figure*}
\centering
\begin{tabular}{p{0.03cm}p{0.03cm}c@{\hspace{0.2em}}c@{\hspace{0.2em}}c@{\hspace{0.2em}}c@{\hspace{0.2em}}c@{\hspace{0.2em}}|@{\hspace{0.2em}}c@{\hspace{0.2em}}c@{\hspace{0.2em}}c@{\hspace{0.2em}}c@{\hspace{0.2em}}c}
\rotatebox{90}{\small Composite}& 
\rotatebox{90}{\small input image}&
\includegraphics[width=0.09\textwidth]{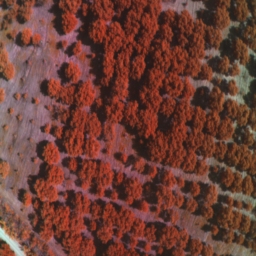}&
\includegraphics[width=0.09\textwidth]{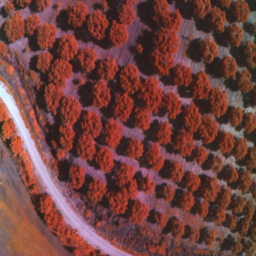}&
\includegraphics[width=0.09\textwidth]{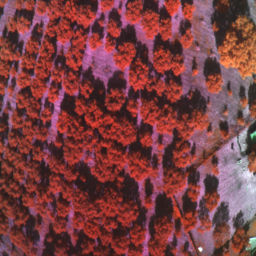}&
\includegraphics[width=0.09\textwidth]{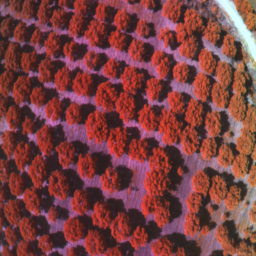}&
\includegraphics[width=0.09\textwidth]{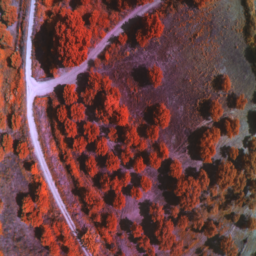}&
\includegraphics[width=0.09\textwidth]{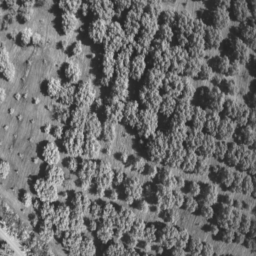}&
\includegraphics[width=0.09\textwidth]{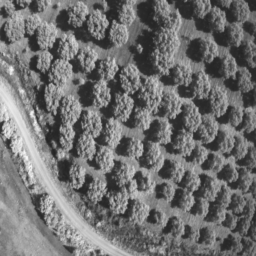}&
\includegraphics[width=0.09\textwidth]{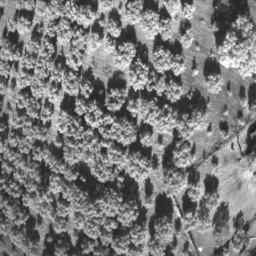}&
\includegraphics[width=0.09\textwidth]{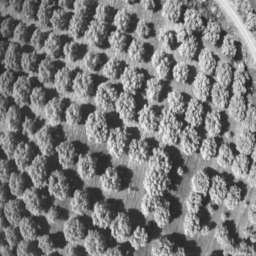}&
\includegraphics[width=0.09\textwidth]{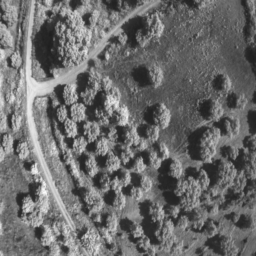}
\\
\rotatebox{90}{\small Groundtruth}& 
\rotatebox{90}{\small image}&
\includegraphics[width=0.09\textwidth]{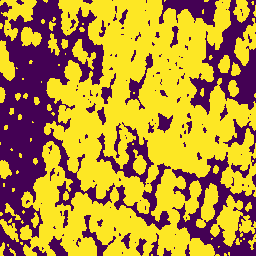}&
\includegraphics[width=0.09\textwidth]{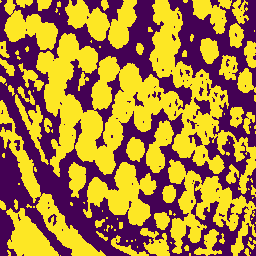}&
\includegraphics[width=0.09\textwidth]{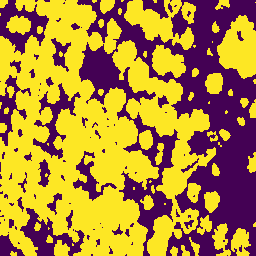}&
\includegraphics[width=0.09\textwidth]{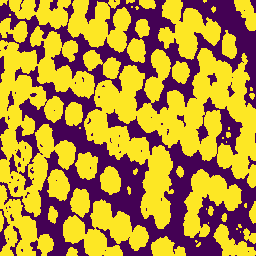}&
\includegraphics[width=0.09\textwidth]{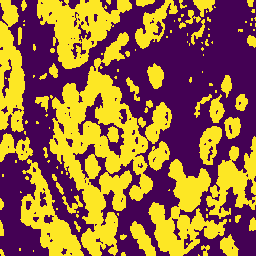}&
\includegraphics[width=0.09\textwidth]{img/label/label1.png}&
\includegraphics[width=0.09\textwidth]{img/label/label2.png}&
\includegraphics[width=0.09\textwidth]{img/label/label3.png}&
\includegraphics[width=0.09\textwidth]{img/label/label4.png}&
\includegraphics[width=0.09\textwidth]{img/label/label5.png}
\\
\rotatebox{90}{\small Base}& 
\rotatebox{90}{\small model}&
\includegraphics[width=0.09\textwidth]{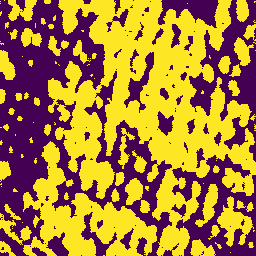}&
\includegraphics[width=0.09\textwidth]{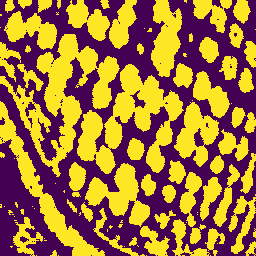}&
\includegraphics[width=0.09\textwidth]{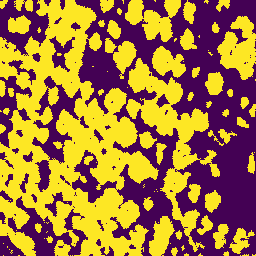}&
\includegraphics[width=0.09\textwidth]{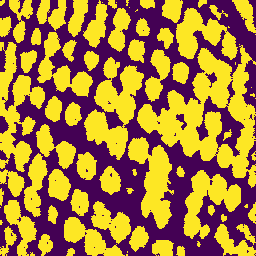}&
\includegraphics[width=0.09\textwidth]{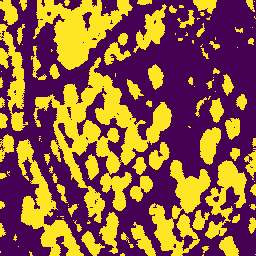}&
\includegraphics[width=0.09\textwidth]{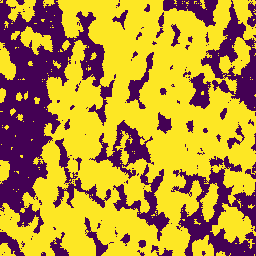}&
\includegraphics[width=0.09\textwidth]{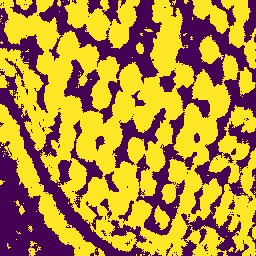}&
\includegraphics[width=0.09\textwidth]{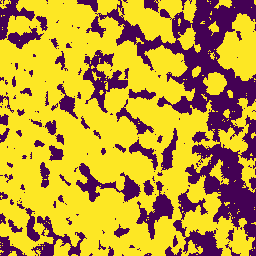}&
\includegraphics[width=0.09\textwidth]{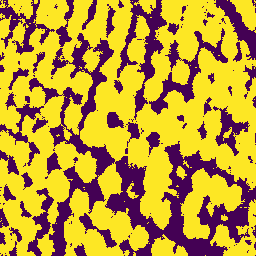}&
\includegraphics[width=0.09\textwidth]{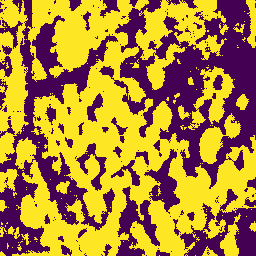}
\\
\rotatebox{90}{\small Quantized}&
\rotatebox{90}{\small model}&
\includegraphics[width=0.09\textwidth]{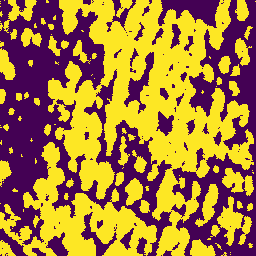}&
\includegraphics[width=0.09\textwidth]{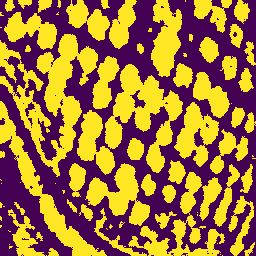}&
\includegraphics[width=0.09\textwidth]{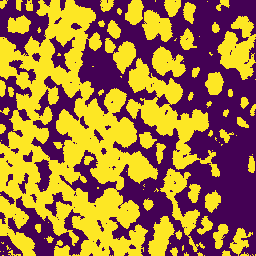}&
\includegraphics[width=0.09\textwidth]{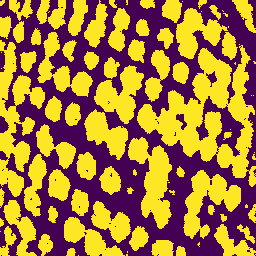}&
\includegraphics[width=0.09\textwidth]{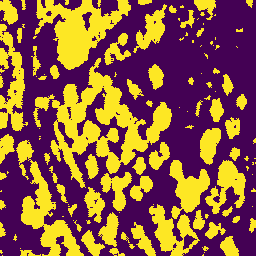}&
\includegraphics[width=0.09\textwidth]{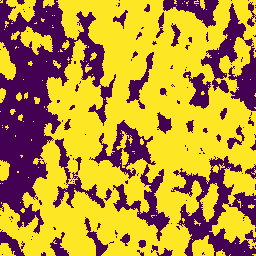}&
\includegraphics[width=0.09\textwidth]{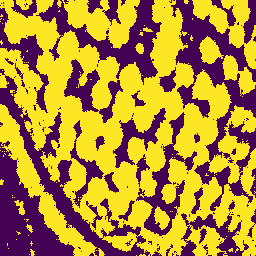}&
\includegraphics[width=0.09\textwidth]{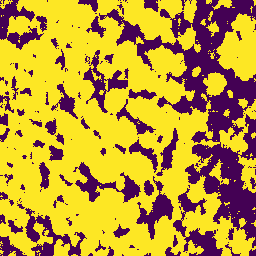}&
\includegraphics[width=0.09\textwidth]{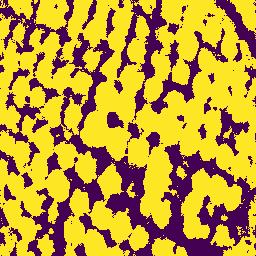}&
\includegraphics[width=0.09\textwidth]{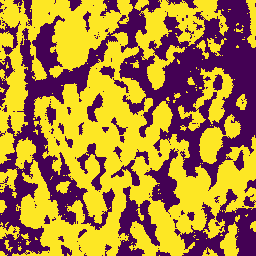}

\\
\end{tabular}
\caption{\small Prediction masks for U-Net trained on multispectral and one-band images. Image input (top); ground truth (second row) and predictions for base model, FlatBuffer model and Quantized model.}
\label{fig:prediction_masks_for_UNet_trained_on_multispectral}
\end{figure*}

\paragraph*{Traditional image processing}
The output of our traditional image processing and watershed based segmentation approach is presented in Figure  \ref{fig:groundtruth_overlay_1}
and Figure~\ref{fig:prediction_masks_for_UNet_trained_on_multispectral}.
By close examination we can derive that tree crowns were successfully segmented from soil and road surrounding (Figures~\ref{fig:groundtruth_overlay_1} and \ref{fig:prediction_masks_for_UNet_trained_on_multispectral}).
The generated pairs of composite or single-band images with the corresponding segmentation maps were used to train the U-Net. 
It is important to highlight that the amount of data required to train the models contains 68 training examples, which is assumed to be a small dataset.

\paragraph*{U-Net segmentation}
For the U-Net based segmentation, a two-step procedure takes place. A preprocessing step prepares the captured image before the U-net extracts the segmentation maps. The preprocessing step for the multispectral-input case involves aligning, cropping and then rescaling the input images, while for the one-band-input case it only involves the rescaling of the input image to the input size of the U-net.

The output of the U-Net segmentation is presented in Figure~\ref{fig:prediction_masks_for_UNet_trained_on_multispectral}.
Qualitative evaluation in terms of visual inspection reveals that the deep architecture adequately segments the tree crowns from the surrounding background. Confusion matrices (Figure~\ref{fig:confusion_multi}) demonstrate an overall accuracy of 89\% for the multi-spectral composite image input and 84\% for the one-band image input. Such observations facilitate the deployment of the trained models on the Google Coral Edge TPU. Performance evaluation in terms of inference times is presented in Table \ref{table:results}. 
Table \ref{tab:execution_time} shows the total execution time for training and inference procedure.
Inference times were calculated on relevant procedures without taking into account I/O operations, generation of composite images
and other configuration calls.
In the case of the base model that meant measuring time spend on the Keras' API call to \texttt{model.predict(image)}. 
This call generates output predictions for the input samples from the loaded model.
In the other models, we measured the time spend to
load the image into the tensor and calling the TFLite Interpreter's method \texttt{interpreter.invoke()}
which is the analogue of Keras' predict for this model format.

Quantized-TPU model executed on Coral Edge requires only 28ms to segment a single image
and is the most efficient inference procedure from those that were considered.
A large spike in inferencing time is observed for the Quantized model when running on PC-CPU;
this is due to the fact that TFLite's routines are unoptimized for x86\_64 architectures.
The Coral Dev Board's CPU has an arm architecture which is supported by TensorflowLite's fully optimized routines.

It is evident that the model trained for multi-spectral input
cannot achieve high performance due to the large
preprocessing overhead of preparing the images for inference.
On the other hand,
using the one-band procedure, 
we can see that
the total execution time is less than 150ms,
when executed on a Coral Edge-TPU,
allowing the segmentation of 7 frames per second.
This can be brought up to 35 frames per second,
by training the U-net for larger one-band image inputs
eliminating, thus, the need to preprocess the images at all.
Furthermore, 
the prediction accuracy in the one-band quantized model 
remains above 80\%,
declining only slightly in comparison to the multi-spectral quantized model.
It is, therefore, feasible to use the one-band procedure
for on-the-fly tree crown delineation,
using only raw one-band input images from UAVs.




\section{Discussion and Conclusion}\label{sec:conclusion}

In this work, we presented a U-Net based tree delineation method. Multi-spectral imagery was captured with drones inspecting olive groves, employing cameras mounted on the chassis. Groundtruth segmentation data were generated using a pipeline of traditional image processing and watershed segmentation approaches. The deep architecture was trained with a relatively small dataset of pairs of multi-spectral one band spectral images and segmentation maps. The trained models were further accelerated with quantization and flatbuffer techniques allowing the execution of the U-Net inference on embedded devices like the Google Coral Edge TPU Board. The experimental evaluation demonstrated that the loss of prediction accuracy was minimal with a significant boost of overall performance, facilitating real-time execution.


\section*{Acknowledgments}

The data were provided by GaiaRobotics Ltd (https://www.gaiarobotics.gr/en/). This paper has received funding from the MyOlivGroveCoach - Analysis, modelling and multi-spectral sensing for the predictive management of Verticillium wilt in olive groves (MIS 5040498) implemented under the Action for the Strategic Development on the Research and Technological Sector, co-financed by national funds through the Operational Programme of Western Greece 2014-2020 and European Union funds (European Regional Development Fund).

\bibliography{output}

\begin{thebibliography}{10}
\providecommand{\url}[1]{#1}
\csname url@samestyle\endcsname
\providecommand{\newblock}{\relax}
\providecommand{\bibinfo}[2]{#2}
\providecommand{\BIBentrySTDinterwordspacing}{\spaceskip=0pt\relax}
\providecommand{\BIBentryALTinterwordstretchfactor}{4}
\providecommand{\BIBentryALTinterwordspacing}{\spaceskip=\fontdimen2\font plus
\BIBentryALTinterwordstretchfactor\fontdimen3\font minus
  \fontdimen4\font\relax}
\providecommand{\BIBforeignlanguage}[2]{{%
\expandafter\ifx\csname l@#1\endcsname\relax
\typeout{** WARNING: IEEEtran.bst: No hyphenation pattern has been}%
\typeout{** loaded for the language `#1'. Using the pattern for}%
\typeout{** the default language instead.}%
\else
\language=\csname l@#1\endcsname
\fi
#2}}
\providecommand{\BIBdecl}{\relax}
\BIBdecl

\bibitem{thorp2004review}
K.~Thorp and L.~Tian, ``A review on remote sensing of weeds in agriculture,''
  \emph{Precision Agriculture}, vol.~5, no.~5, pp. 477--508, 2004.

\bibitem{gougeon1995crown}
F.~A. Gougeon, ``A crown-following approach to the automatic delineation of
  individual tree crowns in high spatial resolution aerial images,''
  \emph{Canadian journal of remote sensing}, vol.~21, no.~3, pp. 274--284,
  1995.

\bibitem{brandtberg1998automated}
T.~Brandtberg and F.~Walter, ``Automated delineation of individual tree crowns
  in high spatial resolution aerial images by multiple-scale analysis,''
  \emph{Machine Vision and Applications}, vol.~11, no.~2, pp. 64--73, 1998.

\bibitem{yang2014multi}
J.~Yang, Y.~He, and J.~Caspersen, ``A multi-band watershed segmentation method
  for individual tree crown delineation from high resolution multispectral
  aerial image,'' in \emph{2014 IEEE Geoscience and Remote Sensing
  Symposium}.\hskip 1em plus 0.5em minus 0.4em\relax IEEE, 2014, pp.
  1588--1591.

\bibitem{li2019real}
W.~Li, C.~He, H.~Fu, J.~Zheng, R.~Dong, M.~Xia, L.~Yu, and W.~Luk, ``A
  real-time tree crown detection approach for large-scale remote sensing images
  on fpgas,'' \emph{Remote Sensing}, vol.~11, no.~9, p. 1025, 2019.

\bibitem{weier2000measuring}
J.~Weier and D.~Herring, ``Measuring vegetation (ndvi \& evi),'' \emph{NASA
  Earth Observatory}, vol.~20, 2000.

\bibitem{zhao2007hierarchical}
K.~Zhao and S.~Popescu, ``Hierarchical watershed segmentation of canopy height
  model for multi-scale forest inventory,'' \emph{Proceedings of the ISPRS
  working group’. pp. 436{\^a}}, vol. 442, 2007.

\bibitem{weinstein2019individual}
B.~G. Weinstein, S.~Marconi, S.~Bohlman, A.~Zare, and E.~White, ``Individual
  tree-crown detection in rgb imagery using semi-supervised deep learning
  neural networks,'' \emph{Remote Sensing}, vol.~11, no.~11, p. 1309, 2019.

\bibitem{huang2018individual}
H.~Huang, X.~Li, and C.~Chen, ``Individual tree crown detection and delineation
  from very-high-resolution uav images based on bias field and
  marker-controlled watershed segmentation algorithms,'' \emph{IEEE Journal of
  Selected Topics in Applied Earth Observations and Remote Sensing}, vol.~11,
  no.~7, pp. 2253--2262, 2018.

\bibitem{santos2019assessment}
A.~A.~d. Santos, J.~Marcato~Junior, M.~S. Ara{\'u}jo, D.~R. Di~Martini, E.~C.
  Tetila, H.~L. Siqueira, C.~Aoki, A.~Eltner, E.~T. Matsubara, H.~Pistori
  \emph{et~al.}, ``Assessment of cnn-based methods for individual tree
  detection on images captured by rgb cameras attached to uavs,''
  \emph{Sensors}, vol.~19, no.~16, p. 3595, 2019.

\bibitem{gomes2018individual}
M.~F. Gomes, P.~Maillard, and H.~Deng, ``Individual tree crown detection in
  sub-meter satellite imagery using marked point processes and a
  geometrical-optical model,'' \emph{Remote Sensing of Environment}, vol. 211,
  pp. 184--195, 2018.

\bibitem{ke2008comparison}
Y.~Ke and L.~J. Quackenbush, ``Comparison of individual tree crown detection
  and delineation methods,'' in \emph{Proceedings of 2008 ASPRS annual
  conference}, 2008.

\bibitem{ke2011review}
------, ``A review of methods for automatic individual tree-crown detection and
  delineation from passive remote sensing,'' \emph{International Journal of
  Remote Sensing}, vol.~32, no.~17, pp. 4725--4747, 2011.

\bibitem{ke2010active}
Y.~Ke, W.~Zhang, and L.~J. Quackenbush, ``Active contour and hill climbing for
  tree crown detection and delineation,'' \emph{Photogrammetric Engineering \&
  Remote Sensing}, vol.~76, no.~10, pp. 1169--1181, 2010.

\bibitem{lin2011multi}
C.~Lin, G.~Thomson, C.-S. Lo, and M.-S. Yang, ``A multi-level morphological
  active contour algorithm for delineating tree crowns in mountainous forest,''
  \emph{Photogrammetric Engineering \& Remote Sensing}, vol.~77, no.~3, pp.
  241--249, 2011.

\bibitem{zhen2015agent}
Z.~Zhen, L.~J. Quackenbush, S.~V. Stehman, and L.~Zhang, ``Agent-based region
  growing for individual tree crown delineation from airborne laser scanning
  (als) data,'' \emph{International Journal of Remote Sensing}, vol.~36, no.~7,
  pp. 1965--1993, 2015.

\bibitem{dalponte2019individual}
M.~Dalponte, L.~Frizzera, and D.~Gianelle, ``Individual tree crown delineation
  and tree species classification with hyperspectral and lidar data,''
  \emph{PeerJ}, vol.~6, p. e6227, 2019.

\bibitem{dai2018new}
W.~Dai, B.~Yang, Z.~Dong, and A.~Shaker, ``A new method for 3d individual tree
  extraction using multispectral airborne lidar point clouds,'' \emph{ISPRS
  journal of photogrammetry and remote sensing}, vol. 144, pp. 400--411, 2018.

\bibitem{jing2012individual}
L.~Jing, B.~Hu, T.~Noland, and J.~Li, ``An individual tree crown delineation
  method based on multi-scale segmentation of imagery,'' \emph{ISPRS Journal of
  Photogrammetry and Remote Sensing}, vol.~70, pp. 88--98, 2012.

\bibitem{duncanson2014efficient}
L.~Duncanson, B.~Cook, G.~Hurtt, and R.~Dubayah, ``An efficient, multi-layered
  crown delineation algorithm for mapping individual tree structure across
  multiple ecosystems,'' \emph{Remote Sensing of Environment}, vol. 154, pp.
  378--386, 2014.

\bibitem{aubry2019comparative}
M.~Aubry-Kientz, R.~Dutrieux, A.~Ferraz, S.~Saatchi, H.~Hamraz, J.~Williams,
  D.~Coomes, A.~Piboule, and G.~Vincent, ``A comparative assessment of the
  performance of individual tree crowns delineation algorithms from als data in
  tropical forests,'' \emph{Remote Sensing}, vol.~11, no.~9, p. 1086, 2019.

\bibitem{csillik2018identification}
O.~Csillik, J.~Cherbini, R.~Johnson, A.~Lyons, and M.~Kelly, ``Identification
  of citrus trees from unmanned aerial vehicle imagery using convolutional
  neural networks,'' \emph{Drones}, vol.~2, no.~4, p.~39, 2018.

\bibitem{li2017deep}
W.~Li, H.~Fu, L.~Yu, and A.~Cracknell, ``Deep learning based oil palm tree
  detection and counting for high-resolution remote sensing images,''
  \emph{Remote Sensing}, vol.~9, no.~1, p.~22, 2017.

\bibitem{ronneberger2015u}
O.~Ronneberger, P.~Fischer, and T.~Brox, ``U-net: Convolutional networks for
  biomedical image segmentation,'' in \emph{International Conference on Medical
  image computing and computer-assisted intervention}.\hskip 1em plus 0.5em
  minus 0.4em\relax Springer, 2015, pp. 234--241.

\bibitem{Hartley2004}
R.~I. Hartley and A.~Zisserman, \emph{Multiple View Geometry in Computer
  Vision}, 2nd~ed.\hskip 1em plus 0.5em minus 0.4em\relax Cambridge University
  Press, ISBN: 0521540518, 2004.

\bibitem{grau2004improved}
V.~Grau, A.~Mewes, M.~Alcaniz, R.~Kikinis, and S.~K. Warfield, ``Improved
  watershed transform for medical image segmentation using prior information,''
  \emph{IEEE transactions on medical imaging}, vol.~23, no.~4, pp. 447--458,
  2004.

\end{thebibliography}
\bibliographystyle{IEEEtran}

\end{document}